\documentclass[%
 reprint,
 amsmath,amssymb,
 aps,
 longbibliography
]{revtex4-2}
\usepackage{hyperref}
\usepackage{graphicx}
\usepackage{dcolumn}
\usepackage{float}
\usepackage{bm}
\usepackage{amsthm}
\usepackage[english]{babel}

\theoremstyle{definition}
\newtheorem*{definition}{Definition}

\usepackage[colorinlistoftodos,prependcaption,textsize=tiny]{todonotes}

\begin{document}

\preprint{APS/123-QED}

\title{Higher order definition of causality by optimally conditioned transfer entropy}

\author{Jakub Ko\v{r}enek$^{1,2}$}
\author{Pavel Sanda$^{1}$}
\author{Jaroslav Hlinka$^{1,3}$}
 \email{hlinka@cs.cas.cz}
\affiliation{$^1$Institute of Computer Science of the Czech Academy of Sciences, Pod Vod\'{a}renskou v\v{e}\v{z}\'{i} 271/2, 182 07 Prague, Czech Republic}
\affiliation{$^2$Faculty of Nuclear Sciences and Physical Engineering, Czech Technical University, B\v{r}ehová 7, 115 19, Prague, Czech Republic}
\affiliation{$^3$National Institute of Mental Health, Topolov\'{a} 748, 250 67 Klecany, Czech Republic
}

\date{\today}

\begin{abstract}
The description of the dynamics of complex systems, in particular the capture of the interaction structure and causal relationships between elements of the system, is one of the central questions of interdisciplinary research. While the characterization of pairwise causal interactions is a relatively ripe field with established theoretical concepts and the current focus is on technical issues of their efficient estimation, it turns out that the standard concepts such as Granger causality or transfer entropy may not faithfully reflect possible synergies or interactions of higher orders, phenomena highly relevant for many real-world complex systems. In this paper, we propose a generalization and refinement of the information-theoretic approach to causal inference, enabling the description of truly multivariate, rather than multiple pairwise, causal interactions, and moving thus from causal networks to causal hypernetworks. In particular, while keeping the ability to control for mediating variables or common causes, in case of purely synergistic interactions such as the exclusive disjunction, it ascribes the causal role to the multivariate causal set but \emph{not} to individual inputs, distinguishing it thus from the case of e.g. two additive univariate causes. We demonstrate this concept by application to illustrative theoretical examples as well as a biophysically realistic simulation of biological neuronal dynamics recently reported to employ synergistic computations.
\end{abstract}

\maketitle


\section{\label{sec:Introduction}Introduction}

The study of complex networks is a rapidly developing field with applications across various scientific disciplines such as neuroscience, climate research, computer science, economics, energetics, or game theory~\cite{Boccaletti2006}. The general approach views a given system as a network of interacting subsystems. A central challenge is to estimate the pattern of interactions from observed data. The formal definition and methods for estimating causal effects from one element to another have been thoroughly studied. 

A common approach is the Granger causality~\cite{Granger1969} - a concept based on two principles: the cause happens before its effect, and the cause carries some additional information about the effect (not included in the 'rest of the universe'). A nonlinear generalization of this concept - transfer entropy~\cite{Schreiber2000, Hlavackova2007} - is based on the same principles. While Granger causality is typically cast in the framework based on prediction via linear vector autoregressive processes, transfer entropy is an information-theoretic measure aiming to capture time-directed information transfer of arbitrary functional form. Indeed, for Gaussian processes, the two concepts are equivalent~\cite{Barnett2009}.

In this work, we are proposing an extension to the framework of Granger causality (seen as reduction in variance) or Transfer Entropy (seen as reduction in surprise). To understand the nature and motivation of the extension, it is elucidating to start by conceptualizing Granger causality as a fundamental refinement of the notoriously naive concept of causality as correlation by adding the requirement of the candidate cause holding correlation (prediction/information) \emph{on top of} that included in the target past and the \emph{rest of the universe}. Note that analogously Transfer Entropy is a refinement of Mutual information by conditioning on target past (and potentially other sources). This key refinement allows ascribing the causal effect more conservatively, in particular controlling for common sources or mediating variables, however is not free of interpretation problems. To alleviate these, we  suggest yet another refinement that would in particular avoid ascribing causal status to variables, that \emph{only} contribute through multivariate interactions. We aim to do this while keeping the original safeguards (against false causal inferences) of Granger's approach .


Let us consider a pristine example of purely \emph{higher order interaction} between source variables in causing the target variables: two candidate source variables $X_1, X_2\stackrel{iid}{\sim} Be(0.5)$, i.e. independent, each with a uniform Bernoulli distribution with probability $p=0.5$; and a target variable $Y$ defined as $Y=XOR(X_1, X_2),$
or in the context of time series $Y(t+1)=XOR(X_1(t),X_2(t)).$ Note that logical XOR is zero if the inputs are equal; otherwise, it is one. In this system, pairwise mutual information  $I(X_1, Y)$  between $X_1$ and $Y$ is equal to zero, and so is $I(X_2, Y)$. In other words, neither of the candidate source variables carries (on their own) information about the target variable $Y$; they play a role, but only together. 
Nevertheless transfer entropy  ascribes causal status to both $X_1$ and $X_2$ (by conditioning each one on the other). One may, however, argue that such representation by a graph with the causal links (from $X_1$ to $Y$ and from $X_2$ to $Y$) is obfuscating something important, as it does not distinguish this system from, say, $Y=X_1+X_2$, where each variable indeed carries information about $Y$ on its own. We suggest a refinement in interpreting the relationship between \( X_1 \) (or \( X_2 \)) and \( Y \): they are, in fact, independent, and thus it may not be appropriate to discuss a direct causal effect in this context. At the same time, representing the system by an empty graph and denying causal status to both variables seems incorrect as well, as that would collide with the case of completely unpredictable $Y$. 
Thus, as only the \emph{multivariate} information $I(\lbrace X_1, X_2 \rbrace, Y)$ is non-zero (in particular 1 bit), we suggest to state that there is no causal link, but there is a \emph{causal hyperlink} from the set $\lbrace X_1, X_2 \rbrace$ to $Y$. Consequently, this perspective necessitates a shift from representing the system using directed graphs to employing directed \emph{causal hypergraphs}. Note that directed hypergraph is a natural generalization of a directed graph, and corresponds to a pair  $ (V,E)$, where  $ V$ is a set of elements called nodes and $ E$ is a set of pairs of subsets of $ V$. Each of these pairs $ (D,C)\in E$ is called a hyperedge; the vertex subset $ D$ is known as its tail, and $ C$ as its head. Less formally one can speak of hypernetworks with hyperlinks between (sets of) nodes, in a causal context from sources to targets. However, note that interpreting the graph-theoretical properties of
"networks" inferred from statistical dependencies is in many situations  problematic and generally a challenging enterprise; see~\cite{Hlinka2017} as an example.

While the XOR function may seem artificial, higher-order interactions are commonly discussed in the context of modelling complex systems
in physics~\cite{Battiston2021}, neuroscience~\cite{Schneidman2006, Shan2011}, ecology~\cite{Mayfield2017}, social sciences~\cite{Cancetti2021} and many more. Despite this challenge is increasingly recognized in the causal inference community~\cite{Javidian2020, Korenek2020,Kugiumtzis2024}, formal refinement of multivariate causality is not yet fully resolved. 



\section{Transfer entropy \& Problem statement}
In the following, we assume that for each target variable $Y$, a set of candidate source variables $\mathbf{X}=\lbrace X_1,\ldots, X_n \rbrace$ is known \emph{a priori}. In practice, the assumption is justified either by Granger causality-like reasoning: the candidate source variables are those temporally preceding the target variable, or by (additional) theoretical arguments. 
The transfer entropy (the positivity of which defines the presence of causal effect from $X_i$ to $Y$) is defined as the conditional mutual information between $X_i$ and $Y$ conditioned on all other potential sources:

\begin{align}
\label{eq:TE}
       TE_{X_i \rightarrow Y}=I\left(X_{i},Y| \mathbf{X} \smallsetminus\lbrace X_{i} \rbrace \right).
\end{align}

As already mentioned, we aim to generalize the transfer entropy concept to more faithfully represent causal structure in the presence of higher-order causal interactions.
A natural generalization is to define existence of a causal effect of a set $\mathbf{X}_I=\lbrace X_{i_1},\ldots, X_{i_k} \rbrace$ on target $Y$ as:

\begin{equation}
    I\left(\lbrace X_{i_1},\ldots, X_{i_k} \rbrace,Y| \mathbf{X} \smallsetminus\lbrace X_{i_1},\ldots, X_{i_k} \rbrace\right)>0.
\end{equation}

Let us revisit the TE behaviour in the XOR example:

\begin{align}
\begin{split}
&X_1, X_2 \stackrel{iid}{\sim} Be(0.5)\\
\label{XOR_eq}
&Y=XOR\left(X_1,X_2\right).
\end{split}
\end{align}

What would such generalization of TE conclude? First, it infers a causal link from the set $\lbrace X_1, X_2 \rbrace$: $I(\lbrace X_1, X_2\rbrace, Y)=1$ bit. However, it also infers causal link from $X_1$ (or $X_2$), as: 
\begin{align}
    \label{chain_rule}I\left(X_1,Y|X_2\right)=I\left(\lbrace X_1,X_2\rbrace,Y\right)-I\left(X_2,Y\right)=1\text{bit}.
\end{align}

As a side note, in practice, these individual links might remain undetected, as commonly used causal network inference algorithms progress iteratively from testing univariate predictors and may stop before discovering such conditional dependence -- see~\cite{Korenek2020} for discussion, and~\cite{Kugiumtzis2024} for an example of an updated (yet, therefore, slower) algorithm more robust in this regard. Also note that the XOR example is straightforwardly generalizable to the k-variable higher order interaction by considering e.g. $Y$ as a sum of variables $X_1, X_2, \ldots X_{k-1} \stackrel{iid}{\sim} Be(0.5)$ modulo 2, i.e. the parity function. Again, even the knowledge of all but one of the source variables provides no information about $Y$ without knowing the last one; it is the interaction of all $k-1$ sources that provides the target predictability. 

Setting aside the implementation details, even such multivariate TE definition does not distinguish the distinctive synergistic causal effect in the XOR example, compared to, e.g. the case of summing two independent normally distributed variables. Two issues arise: first, that TE concludes that $X_1$ has a  'causal effect' on $Y$ although $X_1$ does not hold any information about $Y,$ might be considered superfluous, as the information 'belongs' to the pair (which multivariate TE correctly marks). Second, for the sum function, as the pair does not carry information on top of each variable, one might not want to mark the pair as causal per se, but only each variable. 

\section{Causal effect}
The first issue suggests the suitability of including an extra condition (of non-zero mutual information between the source and the target) in the definition of a causal effect. Indeed, mutual information alone cannot be considered a measure of any causality since it could only mean that variables are affected by a common source, however, it makes good sense to consider it as an additional necessary condition of a (direct) causal effect. 
Such a necessary condition would solve the problem with the above-mentioned example (\ref{XOR_eq}), but the situation can be much more complicated. Consider the following system:
$X_1, \ldots X_p,\mathcal{E}_{p+1},\ldots, \mathcal{E}_{n}\stackrel{iid}{\sim} Be(0.5),$
\begin{align}
\begin{split}
X_k&=X_1+\mathcal{E}_k, \ \ \ k \in \lbrace p+1,\ldots, n \rbrace\\
Y&=\sum_{k=2}^{p}XOR(X_1,X_k)+\sum_{k=p+1}^n X_k
\end{split}
\label{eq:XORsAndIndir}
\end{align}

\begin{figure}[h!]
\centering
\includegraphics[width=0.7\columnwidth]{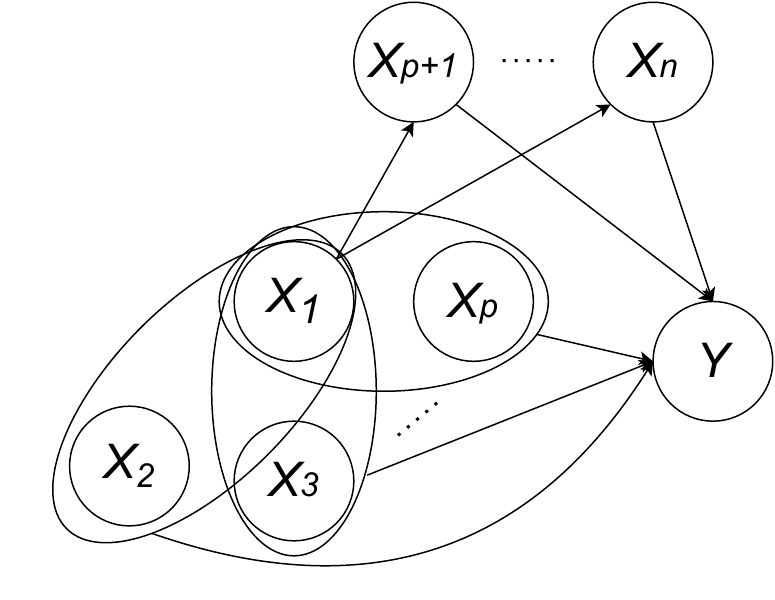}
\caption{Example system~\eqref{eq:XORsAndIndir} illustrating that careful conditioning is necessary to uncover that $X_1$ has only mediated and joint multivariate causal effects on $Y$. Linear terms are represented by oriented edges from a source element to the target element (the element on the left-hand side of the equation), while XOR terms are represented by oriented (hyper)edges from a pair of source elements to the target element.  
}
\label{fig:XORsAndIndir}
\end{figure}

The system is constructed (see Fig.~\ref{fig:XORsAndIndir}) such that $X_1$ does not directly affect $Y$ by itself, but it affects $Y$ indirectly through mediating variables $X_{p+1},\ldots,X_n.$ Also $X_1$ affects $Y$ through pairwise interactions: each pair $\lbrace X_1, X_i \rbrace$ for $ i \in \lbrace 2,\ldots,p \rbrace$ affects $Y$. Note that both the mutual information $I(X_1, Y)$ and the fully conditional mutual information (transfer entropy) $I\left(\left( X_1, Y\right)|\mathbf{X}\smallsetminus \lbrace X_1 \rbrace \right)$ are greater than zero, suggesting thus a causal effect of $X_1$ on $Y$. In fact, all of the conditional mutual informations $I(X_1,Y|\mathbf{S})$ are positive, except $I\left(X_1,Y|X_{p+1},\ldots,X_n\right)$. 

Indeed, if any of variables $\lbrace X_2, \ldots, X_p \rbrace$ is in the condition, the information is non-zero because of the effect of the predictor $\lbrace X_1, X_i \rbrace$ (follows from chain rule - eq.~(\ref{chain_rule})). On the other hand, if any of variables $\lbrace X_{p+1}, \ldots, X_n \rbrace$ is missing in the condition, information is non-zero because they are mediators of $X_1$ and $Y$ (information flows indirectly from $X_1$ to $Y$ through them). Thus, we suggest to define a causal effect as follows.
\begin{definition}[Unaccountable causal effect \& Optimally conditioned causal entropy]
Let $\mathbf{X}=\lbrace X_1,\ldots, X_n \rbrace$ be a set of candidate source variables and $Y$ a target variable. There is a causal effect from set $ \mathbf{X}_I=\lbrace X_{i_1},\ldots, X_{i_k} \rbrace$ to $Y$ if and only if: 
\begin{align}
\label{Causality_Korenek}
    I(\lbrace X_{i_1},\ldots, X_{i_k} \rbrace, Y|\mathbf{S})>0
\end{align}
for all $\mathbf{S} \subseteq \mathbf{X}\smallsetminus \lbrace
X_{i_1},\ldots, X_{i_k} \rbrace.$ 

The \emph{optimally conditioned transfer entropy} is given as:
\begin{align}
    OCTE_{\mathbf{X}_I \rightarrow Y}=\min_{\mathbf{S}\subseteq \mathbf{X} \smallsetminus \mathbf{X}_{I}}I\left(\mathbf{X}_I, Y| \mathbf{S} \right).
\end{align}   
\end{definition}

In words, the causal effect is the minimum information a variable carries about a target, across conditioning it by all possible subsets of the 'rest of the universe' - instead of considering only exactly the 'rest of the universe' in the condition. Informally, a variable is only considered truly causal if it can not be conditioned out, whichever subset of other variables we choose.

Notably, the definition of the causal effect of a set of variables typically enforces the causal influence of an arbitrary set containing a variable with a causal effect. Consider, for example, a simple linear system where the variable $Y$ is causally affected just by variable $X_1$ plus some noise: $Y=X_1+\mathcal{E}_Y$, and in the system, there are also other variables $\lbrace X_2,\ldots, X_n \rbrace$ independent on $Y$ and $X_1$. Then, all sets containing $X_1$ are also causally affecting $Y$. Note that similar behaviour (supersets of causes defined as causes irrespective of further value in prediction) also holds for (a multivariate variant of) transfer entropy. 
In other words, although the definition of causal effect avoids attributing higher-order causal effects to individual variables, it still does not by itself distinguish multivariate causes that are trivial in the sense of inherited from its subsets from nontrivial higher-order multivariate causal effects.

We suggest one may add this distinction of a \emph{unique} higher-order causal effect by interpreting differently the observation of multivariate causes in the presence/absence of the causal effects of the subsets. Further, to distinguish purely additive from synergistic multivariate causality, one may quantify the higher order causal information not induced trivially by the subsets; suitable approaches may include maximum entropy definition of higher order interactions~\cite{Schneidman2006,Martin2016}, or partial information decomposition (PID)~\cite{Beer2010}. The relationship between OCTE and PID is inherently close but non-trivial. For three variables, PID can be constructed using OCTE: the unique information of an individual variable can be defined as its OCTE, while the redundancy and synergy elements are derived from the multivariate information. This construction satisfies the axioms of PID. However, a more natural approach, defining unique information and synergy directly as OCTE (with synergy corresponding to the OCTE of the set of source variables)—fails to meet the PID axioms. A more sophisticated approach to PID, incorporating the strategy of powerset minimization as utilized in the analysis of how environmental factors influence transfer entropy between two variables, was recently proposed in ~\cite{Stramaglia2024}.

Returning to the XOR example, note that the resulting causal graph of the system also depends on the distribution of the variables $X_1$ and $X_2.$ 
Consider three different distributions, see Table \ref{XOR_table}. In all cases $p_A, p_B, p_C$, $X_1$ and $X_2$ are independent variables with Bernoulli distribution.

\begin{table}[h]
\centering
\begin{tabular}{|l|c|c|c|}
\hline
 & $a)$ & $b)$ & $c)$ \\ 
\hline
$p(X_1 = 0, X_2 = 0, Y = 0)$ & 0.25 & 0.10 & 0.06 \\ 
$p(X_1 = 0, X_2 = 1, Y = 1)$ & 0.25 & 0.40 & 0.24 \\ 
$p(X_1 = 1, X_2 = 0, Y = 1)$ & 0.25 & 0.10 & 0.14 \\ 
$p(X_1 = 1, X_2 = 1, Y = 0)$ & 0.25 & 0.40 & 0.56 \\ 
$I(X_1, Y)$ & 0.00 & 0.28 & 0.24 \\ 
$I(X_2, Y)$ & 0.00 & 0.00 & 0.08 \\ 
$I(X_1, Y | X_2)$ & 1.00 & 1.00 & 0.88 \\ 
$I(X_2, Y | X_1)$ & 1.00 & 0.72 & 0.72 \\ 
$I(\{X_1, X_2\}, Y)$ & 1.00 & 1.00 & 0.96 \\ 
\hline
\end{tabular}
\caption{Example infromation-theoretical functionals for a triplet of variables coupled by the function $XOR(X_1,X_2),$ where $X_1$ and $X_2$ are independent random variables with distributions: $i)$ $p_A: \ X_1,X_2 \sim Be(0.5),$ $ii)$ $p_B: \ X_1\sim Be(0.5), \  X_2\sim Be(0.8)$ $iii)$ $p_C: \ X_1\sim Be(0.7), \  X_2\sim Be(0.8).$ Note that the presence of non-zero (conditional) information values depends not only on the coupling function, but substantially on the input variable distributions, even under independence of inputs.}  
\label{XOR_table}
\end{table}



We have already discussed the case A with  uniformly distributed probabilities. In case B, $X_1$ is uniformly distributed, but $X_2\sim Be(0.8).$ While $I(X_2,Y)$ remains zero, $I(X_1,Y)$ is positive, as is $I(X_1,Y|X_2).$ Variable $X_1$ thus has a causal effect on $Y$ only due to the change in the distribution $X_2.$ The further the variable $X_2$ deviates from the uniform distribution, the more information $X_1$ has about $Y$ and the unique contribution of the pair $\lbrace X_1, X_2 \rbrace$ decreases. In case C, where both variables deviate from the uniform distribution, both $X_1$ and $X_2$ have a causal effect on $Y$, and they also have a unique contribution as a pair $\lbrace X_1, X_2 \rbrace.$ 

It is worth mentioning that a similar challenge in defining causal effects, as seen in examples such as the XOR function, also arises in Judea Pearl's approach to causality~\cite{Pearl09}. Pearl's total effect of a variable \( X \) on an outcome \( Y \) is defined as a difference in the probability distribution of \( Y \) when \( X \) is manipulated versus when \( X \) is observed naturally. Specifically, the total effect is determined by comparing the probabilities \( P(Y_x = y) \) and \( P(Y = y) \). Here, \( P(Y_x = y) \) denotes the probability that \( Y \) takes the value \( y \) after an intervention on \( X \) (i.e., when \( X \) is set to a specific value), while \( P(Y = y) \) represents the probability that \( Y \) takes the value \( y \) under unperturbed, observational conditions (without any intervention on \( X \)). If \( P(Y_x = y) \neq P(Y = y) \), this inequality indicates that the intervention on \( X \) has caused a shift in the distribution of \( Y \), suggesting the existence of a total causal effect. 

 Further, the direct causal effect of \( X \) on \( Y \), holding other variables \( \mathbf{Z} \) constant, is defined as the difference \( P(Y_{x\mathbf{z}} = y) - P(Y_{x^*\mathbf{z}} = y) \), where \( Y_{x\mathbf{z}} \) represents the outcome after both \( X \) and \( \mathbf{Z} \) are set to specific values ($X=x$ and $\mathbf{Z}=\mathbf{z}$), and \( Y_{x^*\mathbf{z}} \) represents the outcome after \( X \) is set to \( x^* \) (different from $x$) while \( \mathbf{Z} \) are set equal to the previous case ($\mathbf{Z}=z$)~\cite{Pearl2022}. This definition aims to isolate the direct effect of \( X \) on \( Y \), removing the influence of other variables.

When applying the concept of total causal effect to the example \( Y = XOR(X,Z) \), we observe that \( P(Y_x = y) = \text{Be}(0.5)=P(Y = y) \) for any value of \( x \), interpreted as the absence of total causal effect of \( X \) on \( Y \). However, \( Y_{x=0, z=0} = 0 \) and \( Y_{x=1, z=0} = 1 \), therefore indicating a direct effect of \( X \) on \( Y \).

The absence of total causal effect in the presence of a direct causal effect appears contradictory and difficult to interpret. 
However, we propose one can apply the same strategy as in the case of transfer entropy, in particular by redefining a variable as having a direct causal effect if and only if its perturbations affect the distribution irrespective of which subset of the other system variables is kept fixed.

\section{Dendritic computation of XOR}
Higher order (synergistic) dependencies occur in systems across disciplines, such as computer science or neuroscience. 
For instance, XOR based detectors are frequently used in algorithmic image edge detection~\cite{al2016,diaconu2013}. For each pixel, the logical XOR mask is applied to a pair of adjacent pixels, and positive XOR value marks an edge in the picture. 
A similar processing principle can be found in some retinal ganglion cells, which respond to differences in intensity across the receptive field while ignoring spatially uniform input and make those neurons well posited to detect edges in visual images~\cite{Berry}.

\begin{figure}[h!]
\centering
\includegraphics[width=0.5\textwidth, trim={0 7cm 0 7cm},clip]{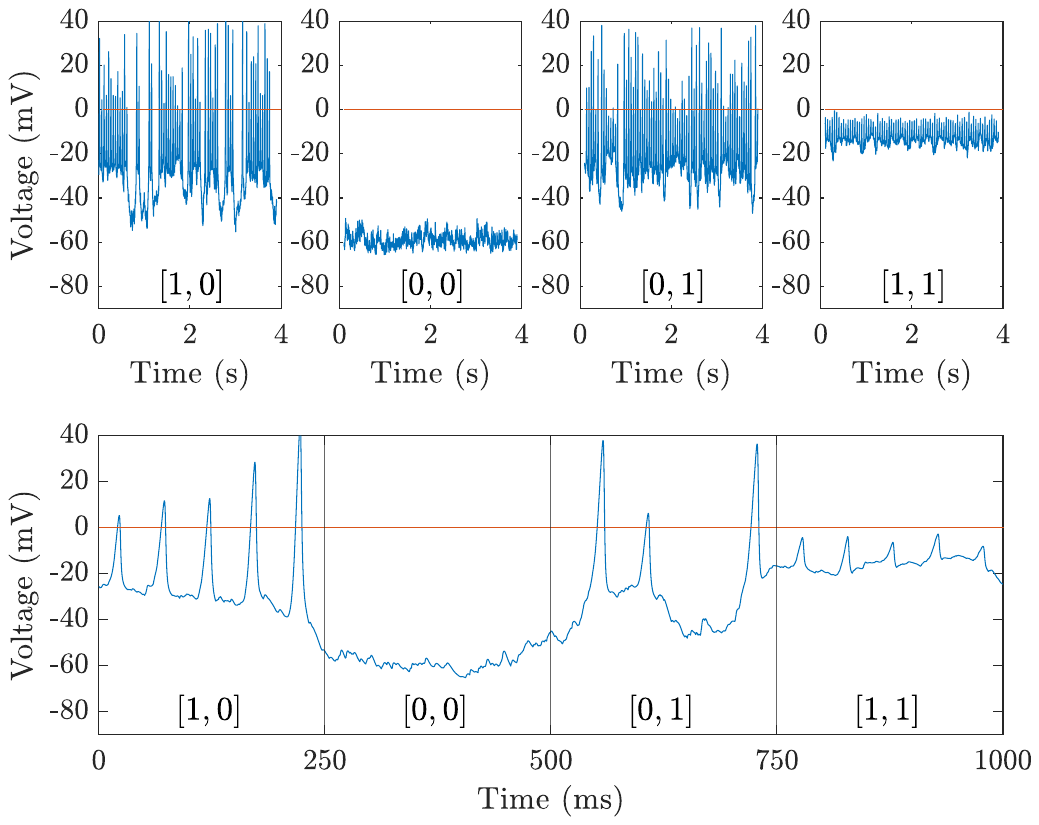}
\caption{Membrane voltage at the dendritic site of the dCaAP mechanism. Top. Dynamics of 4 logical configurations. In each panel, the system receives stable synaptic input for 4~s from two distinct synaptic pathways [$X_1$, $X_2$], the associated logical values shown in labels at the bottom. Unequal input ([0,1] or [1,0]) causes large dendritic spikes. Coincident activity in both pathways ([1,1]) shows as depolarized compared to inactive pathways ([0,0]) but will not make the neuron fire. Bottom. Altering input for each 250~ms window. 
}
\label{fig:Phases}
\end{figure}

We further focus more closely on another example of higher order causal interactions in neuronal dynamics. Despite the long-standing assumption that computing logical XOR requires a circuit of connected neurons, recent work showed that calcium mediated dendritic action potentials (dCaAPs) in a single human cortical neuron can compute this function~\cite{Gidon2020}. In essence, this computation is provided by anti-coincidence behaviour on the apical dendrite of the pyramidal neuron, where two simultaneous synaptic inputs surprisingly reduce dCaAP amplitude (in contrast to the traditional amplification of dendritic AP). We use the biophysical model of dCaAPs provided by~\cite{Gidon2020}, simulating its dynamics while dynamically modulating the synaptic inputs coming from two distinct sources, the activity of which corresponded to two parallel sequentially generated logical inputs, where low/high synaptic input represented logical value of 0/1 respectively. The resulting activity at the apical dendrite ("XOR gate") is shown in Fig.~\ref{fig:Phases}.

\begin{figure}[h!]
\centering
\includegraphics[width=0.5\textwidth, trim={0 7cm 0 7cm},clip]{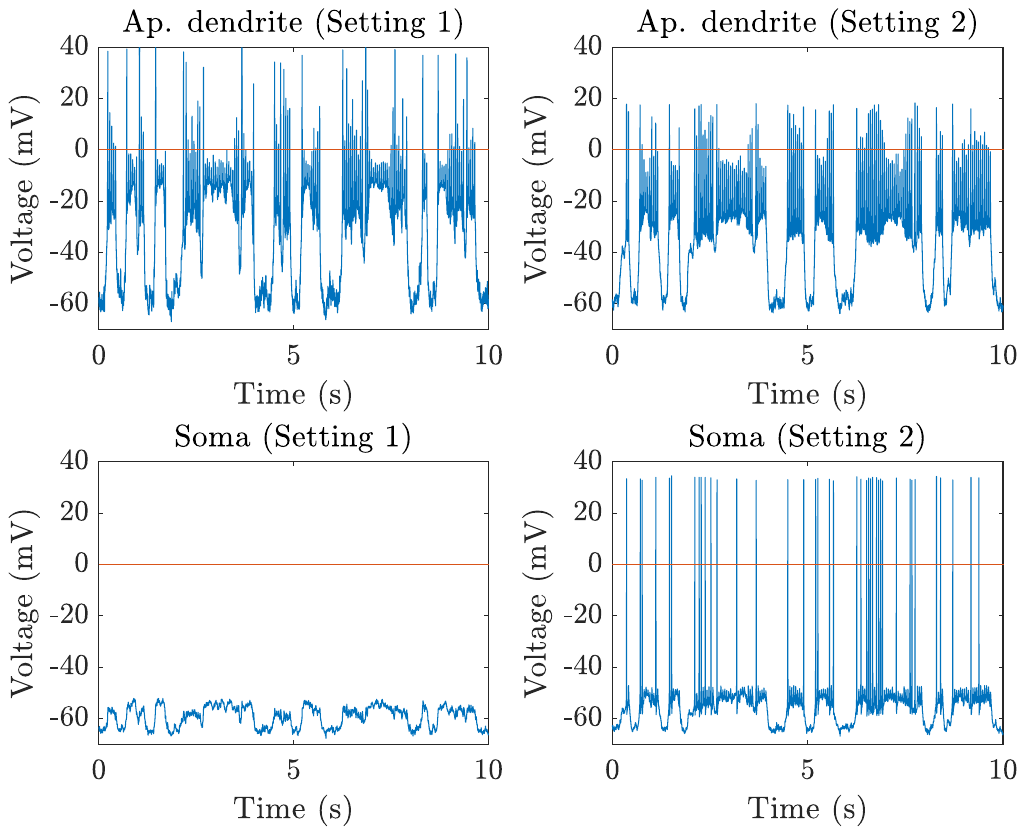}
\caption{Example of membrane voltage on the apical dendrite and soma of the neuron. Top panels. Membrane potential at the dendritic site of dCaAP mechanism (as in Fig.~\ref{fig:Phases}). Bottom. Membrane voltage at the soma of the neuron. Left panels -- distant case. dCaAP mechanism is located 550 $\mu$m from the soma (corresponds to \cite{Gidon2020}, Fig. 3). Right -- proximal case. dCaAP mechanism is located 287 $\mu$m from the soma (corresponds to \cite{Gidon2020}, Fig. S9). For model details, see the code deposition in modelDB (https://modeldb.science/2016664).
}
\label{fig:Voltage}
\end{figure}

Following~\cite{Gidon2020}, we consider two settings: in setting 1, the location of the dCaAP mechanism on the dendrite is distant from the soma (Fig.~\ref{fig:Voltage}, left), and in setting 2 the location is closer, thereby allowing some of the large dendritic spikes to trigger somatic spikes (Fig.~\ref{fig:Voltage}, right).

The simulated 100 seconds were divided into 250~ms windows.  
Then we binarized to the apical dendrite voltages  (Fig.~\ref{fig:Voltage}, top) to obtain the target variable $Y$ ($\tilde{Y}$ for setting 2): we assigned a value of 1 to the $Y$ if a spiking rate in the  window is above 10~Hz and 0 otherwise (spike was defined as the membrane voltage exceeding 0~mV).

\begin{figure}[h!]
\centering
\includegraphics[width=0.45\textwidth, trim={0 11cm 0 11cm},clip]{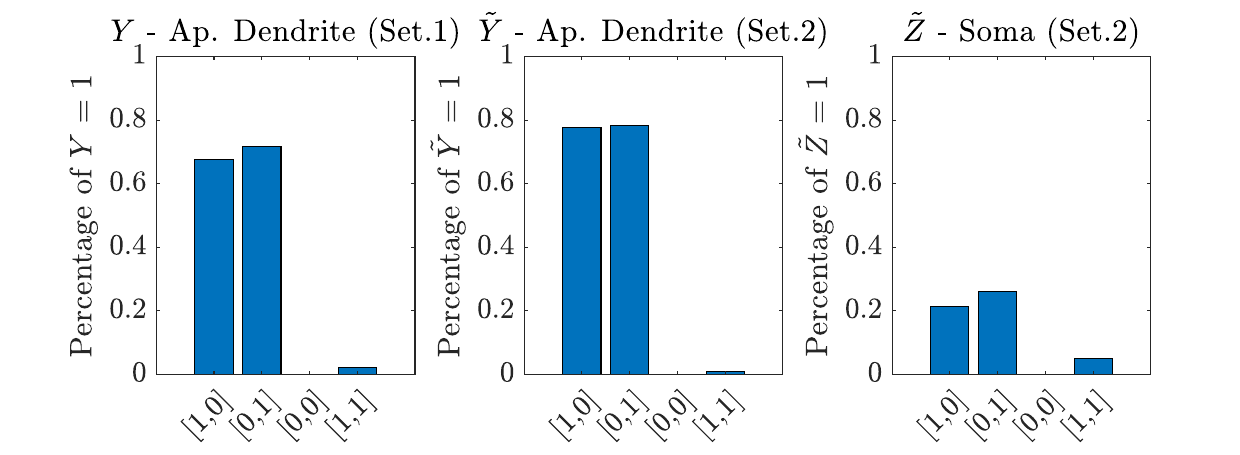}
\caption{Distribution of output variables $Y$, $\tilde{Y}$ and $\tilde{Z}$ depending on input configuration $[X_1, X_2]$.  Left. Distribution on the apical dendrite -- distant case. Middle/Right. Distribution on the apical dendrite/soma -- proximal case.
}
\label{fig:Percentage}
\end{figure}

Fig.~\ref{fig:Percentage} shows the measured distribution of the output variables. The distribution in all three cases follows XOR distribution, albeit for the somatic compartment, it is less accurate due to noisy transmission from the dendrite.

To determine causal relationships, the (conditional) mutual information between the input variables ($X_1,$ $X_2$ and $\lbrace X_1, X_2\rbrace$) and the target variable $Y$ was evaluated. As the finite sample estimate of mutual information is generally nonzero, even for independent variables, a statistical test is required. To evaluate the null hypothesis $I(\mathbf{X}_I,Y|\mathbf{S})=0$ at significance level $\theta=0.01,$ we use a permutation test (using $N=1000$ realizations of shuffled source variable(s) $\mathbf{X}_I$ to generate the null distribution). This breaks any dependency (pairwise or multivariate) between $\mathbf{X}_I$ and $Y$ while maintaining dependencies between $Y$ and $\mathbf{S}$.

For setting 1, we infer both $I(X_1, Y)=0$ and $I(X_2, Y)=0$, while $I(\lbrace X_1, X_2 \rbrace, Y)=0.46$; thus we conclude by the definition of causality (\ref{Causality_Korenek}) there is no causal influence from $X_1$ or $X_2$, but there is a causal (hyper)link from $\lbrace X_1, X_2\rbrace$ to $Y$ with a causal strength of $OCTE_{\lbrace X_1, X_2 \rbrace \rightarrow Y}=0.46$. For the setting 2, we get the same causal structure between the synaptic input and activity on apical dendrite with a causal strength of $OCTE_{\lbrace X_1, X_2 \rbrace \rightarrow \tilde{Y}}=0.56$.

\begin{figure}[h!]
\centering
\includegraphics[width=0.32\textwidth]{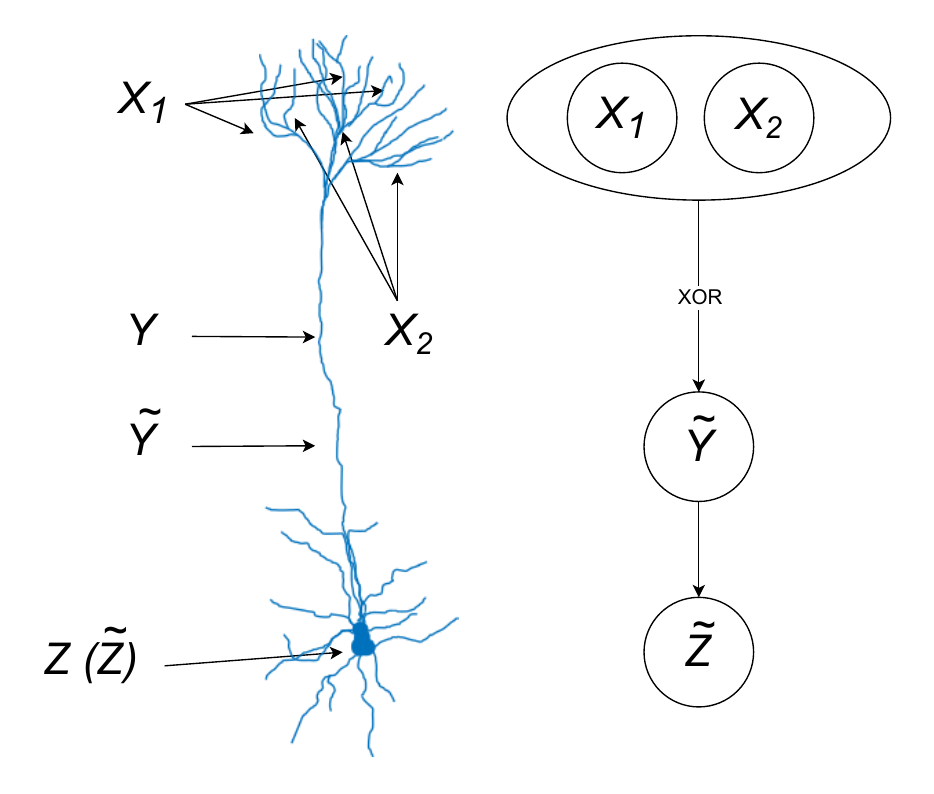}
\caption{Left: Scheme of pyramidal neuron. $X_1$ and $X_2$ mark two distinct synaptic pathways (inputs), $Y$ the output of the dCaAP mechanism on the apical dendrite ($\tilde{Y}$ for setting 2), $Z$ $(\tilde{Z})$ the neuron soma. Right: Corresponding causal diagram.}
\label{fig:neuron}
\end{figure}

We can further evaluate the mediating role of the apical dendrite on the somatic spiking in setting 2. We add the variable $\tilde{Z}$ representing somatic spiking to the monitored system (see right bottom corner of the Fig.~\ref{fig:Voltage}). 
The set of potential sources for target variable $\tilde{Z}$ are all subsets of $\lbrace X_1, X_2, Y \rbrace$. We obtain $I(X_1, \tilde{Z})=0$, $I(X_2, \tilde{Z})=0,$ thus neither $X_1$ nor $X_2$ are causal parents. Although  $I(\lbrace X_1, X_2 \rbrace, \tilde{Z})\neq 0$, the predictor $\lbrace X_1, X_2 \rbrace$ is not a causal parent of $ \tilde{Z}$ because $I(\lbrace X_1, X_2 \rbrace, \tilde{Z}| \tilde{Y})=0.$ In this case, all information is mediated by $\tilde{Y},$ thus link from $\lbrace X_1, X_2 \rbrace$ to $\tilde{Z}$ is indirect. The only nontrivial causal parent of the $\tilde{Z}$ variable is $\tilde{Y}$ with causal strength $OCTE_{\tilde{Y} \rightarrow \tilde{Z}}=0.04$ (as discussed, all the supersets of $\tilde{Y}$ also are causal parents of $\tilde{Z}$, however, as here they carry no extra information, being thus a purely formal, rather than unique, multivariate cause). The estimated causal scheme, a sketch of the neuron, is shown in Figure~\ref{fig:neuron}.

\vspace{0.1pt}

\section{Conclusions}
We studied the problem of defining the causal effect in systems exhibiting higher-order causal patterns such as XOR-like interactions. We showed that the traditional approaches do not provide sufficiently refined answer. We tackle this challenge by introducing a new, refined definition of causal structure and discuss its relation to the original definition. A trade-off for the increased expressiveness of the causal hypernetwork definition is its exponential complexity due to formally requiring iteration over a powerset in the condition. This calls for iterative approximation algorithms already commonly utilized for transfer entropy estimation. The main theoretical challenge lies then in treating conveniently the cases of combined synergistic and direct, or even mediating, causal effect.

\begin{acknowledgments}
This work was supported by GACR 21-17211S, ERDF-Project Brain dynamics, No. CZ.02.01.01/00/22\_ 008/0004643 and SGS23/187/OHK4/3T/14.
\end{acknowledgments}

\bibliographystyle{apsrev4-2}
\bibliography{biblio}

\end{document}